\documentclass[graybox]{svmult}

\usepackage{mathptmx}       
\usepackage{helvet}         
\usepackage{courier}        
\usepackage{type1cm}        
\usepackage{makeidx}         
\usepackage{graphicx}        
\usepackage{multicol}        
\usepackage[bottom]{footmisc}

\makeindex             

\usepackage{amssymb}
\usepackage{pifont}
\usepackage{multirow}
\usepackage{subcaption}
\usepackage{amsmath}

\DeclareMathAlphabet{\mathcal}{OMS}{cmsy}{m}{n}

\newcommand{\csQualitative}[2]{
	\includegraphics[width=#1\textwidth]{img/paper_images_addendum/cityscapes_imgs/gt/#2} &
	\includegraphics[width=#1\textwidth]{img/paper_images_addendum/cityscapes_imgs/exp4_selection/inpaint_selection/#2} &
	\includegraphics[width=#1\textwidth]{img/paper_images_addendum/cityscapes_imgs/exp4_selection/nn/#2} \\
}

\newcommand{\MICCsriQualitativeBaselines}[2]{
\scriptsize	Input & \scriptsize Ground Truth & \scriptsize Ours \\
	\includegraphics[width=#1\textwidth]{img/paper_images_addendum/carla_imgs/gt/#2} &
	\includegraphics[width=#1\textwidth]{img/paper_images_addendum/carla_imgs/gt_static/#2} &
	\includegraphics[width=#1\textwidth]{img/paper_images_addendum/carla_imgs/inpaint/#2} \\
\scriptsize	Nearest Neighbor & \scriptsize Navier-Stokes \cite{bertalmio2001navier} & \scriptsize PatchMatch \cite{barnes2009patchmatch} \\
	\includegraphics[width=#1\textwidth]{img/paper_images_addendum/carla_imgs/nn/#2} &
	\includegraphics[width=#1\textwidth]{img/paper_images_addendum/carla_imgs/navierstokes/#2} &
	\includegraphics[width=#1\textwidth]{img/paper_images_addendum/carla_imgs/photoshop/#2}  \\ \vspace{-5pt} \\
	
}

\newcommand{\MICCexpTwo}[2]{
\scriptsize	Input & \scriptsize Ground Truth & \scriptsize Ours \\
	\includegraphics[width=#1\textwidth]{img/paper_images_addendum/carla_imgs/exp2/carla_deeplab_dynamic_finetuned_colored/#2} &
	\includegraphics[width=#1\textwidth]{img/paper_images_addendum/carla_imgs/exp2/carla_deeplab_static_finetuned_colored/#2} &
	\includegraphics[width=#1\textwidth]{img/paper_images_addendum/carla_imgs/exp2/inpainting/#2} \\
\scriptsize	Nearest Neighbor & \scriptsize Navier-Stokes \cite{bertalmio2001navier} & \scriptsize PatchMatch \cite{barnes2009patchmatch} \\
	\includegraphics[width=#1\textwidth]{img/paper_images_addendum/carla_imgs/exp2/nn/#2} &
	\includegraphics[width=#1\textwidth]{img/paper_images_addendum/carla_imgs/exp2/navier/#2} &
	\includegraphics[width=#1\textwidth]{img/paper_images_addendum/carla_imgs/exp2/carla_deeplab_dynamic_finetuned_colored_patchmatch/#2}  \\ \vspace{-5pt} \\
}

\newcommand{\MICCexpThree}[2]{
\scriptsize	Input & \scriptsize Ground Truth & \scriptsize Ours \\
	\includegraphics[width=#1\textwidth]{img/paper_images_addendum/carla_imgs/exp3/RGB_dynamic_resized/#2} &
	\includegraphics[width=#1\textwidth]{img/paper_images_addendum/carla_imgs/exp2/carla_deeplab_static_finetuned_colored/#2} &
	\includegraphics[width=#1\textwidth]{img/paper_images_addendum/carla_imgs/exp3/deeplabbed_rgb_generative_inpainting/#2} \\
\scriptsize	Nearest Neighbor & \scriptsize Navier-Stokes \cite{bertalmio2001navier} & \scriptsize PatchMatch \cite{barnes2009patchmatch} \\
	\includegraphics[width=#1\textwidth]{img/paper_images_addendum/carla_imgs/exp3/DEEPLABBED_nn_on_DEEPLAB_rgb_colored/#2} &
	\includegraphics[width=#1\textwidth]{img/paper_images_addendum/carla_imgs/exp3/selection_navier_deeplabbed_FIX_colored/#2} &
	\includegraphics[width=#1\textwidth]{img/paper_images_addendum/carla_imgs/exp3/carla_rgb_deeplab_patchmatch_colored_quantized/#2}  \\ \vspace{-5pt} \\
}

\begin{document}

\title*{Road layout understanding by generative adversarial inpainting}
\author{Lorenzo Berlincioni, Federico Becattini, Leonardo Galteri, Lorenzo Seidenari, Alberto Del Bimbo}
\authorrunning{L. Berlincioni, F. Becattini, L. Galteri, L. Seidenari, A. Del Bimbo}
\institute{Lorenzo Berlincioni, Federico Becattini, Leonardo Galteri, Lorenzo Seidenari, Alberto Del Bimbo
\and MICC - University of Florence \and name.surname@unifi.it}
%
%
\maketitle

\abstract*{Autonomous driving is becoming a reality, yet vehicles still need to rely on complex sensor fusion to understand the scene they act in.
The ability to discern static environment and dynamic entities provides a comprehension of the road layout that poses constraints to the reasoning process about moving objects.
We pursue this through a GAN-based semantic segmentation inpainting model to remove all dynamic objects from the scene and focus on understanding its static components such as streets, sidewalks and buildings. We evaluate this task on the Cityscapes dataset and on a novel synthetically generated dataset obtained with the CARLA simulator and specifically designed to quantitatively evaluate semantic segmentation inpaintings. We compare our methods with a variety of baselines working both in the RGB and segmentation domains.}

\abstract{Autonomous driving is becoming a reality, yet vehicles still need to rely on complex sensor fusion to understand the scene they act in.
The ability to discern static environment and dynamic entities provides a comprehension of the road layout that poses constraints to the reasoning process about moving objects.
We pursue this through a GAN-based semantic segmentation inpainting model to remove all dynamic objects from the scene and focus on understanding its static components such as streets, sidewalks and buildings. We evaluate this task on the Cityscapes dataset and on a novel synthetically generated dataset obtained with the CARLA simulator and specifically designed to quantitatively evaluate semantic segmentation inpaintings. We compare our methods with a variety of baselines working both in the RGB and segmentation domains.
}

\begin{figure*}[t]
	\centering
	\begin{tabular}{ccc}
		\includegraphics[width=0.3\textwidth]{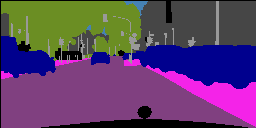} &
		\includegraphics[width=0.3\textwidth]{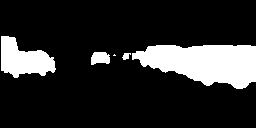} &
		\includegraphics[width=0.3\textwidth]{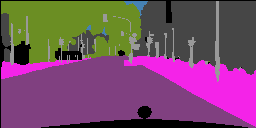} \\
		\includegraphics[width=0.3\textwidth]{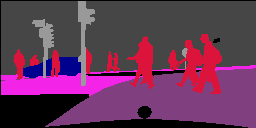} &
		\includegraphics[width=0.3\textwidth]{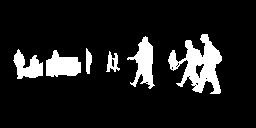} &
		\includegraphics[width=0.3\textwidth]{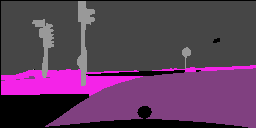}\\
		\includegraphics[width=0.3\textwidth]{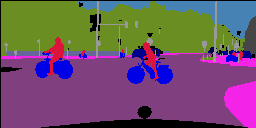} &
		\includegraphics[width=0.3\textwidth]{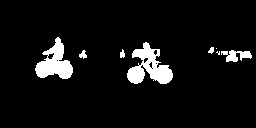} &
		\includegraphics[width=0.3\textwidth]{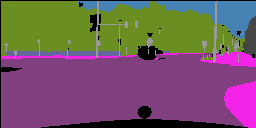}
	\end{tabular}
	\caption{Input and output segmentations of our method along with the inpainting mask of the dynamic object to remove. Left: input. Center: dynamic objects mask. Right: inpainted segmentation.}
	\label{img:inpainting}
\end{figure*}

\section{Introduction and Related Work}
\label{sec:intro}

Autonomous cars rely on a plethora of sensors to understand the environment they move in. Several cues are fused to feed the decision process leading to path planning comprising obstacle avoidance and emergency maneuvers. While depth is often acquired through the non exclusive combination of stereo vision and active sensors, the main cue to compute free space, predict danger and plan future commands is a pixel-wise semantic map. Such maps are usually extracted combining pixel level predictions~\cite{chen2018deeplab} with instance based segmentations~\cite{kirillov2018panoptic}, depth may be fused to enhance the performance~\cite{wang2018depth, uhrig2016pixel}.

Recently, image generation has become an important component of autonomous vehicle system development. Generating images allows to avoid costly acquisitions and possibly the simulation of unlikely but relevant events. Images generation is a pixel-wise operation in which an unseen image is created from a source. Source and target domains may be the same, e.g. $RGB \rightarrow RGB$ or not $\mathcal{C} \rightarrow RGB$, where $\mathcal{C}$ indicates the pixel semantic label. Datasets are often acquired in certain lighting and weather conditions. Simulating a different weather or time of day for the same sequences gives access to a wider set of training samples~\cite{CycleGAN2017}. In certain cases games have been used for such task~\cite{richter2016playing}, the advantage of generating synthetic images through a 3D engine is the ready and precise availability of semantic and depth ground truth data.

A variety of computer vision applications can be seen as image-to-image translation problems between two domains~\cite{isola2017image, CycleGAN2017, xie2015holistically, pathak2016context, long2015fully, eigen2015predicting, johnson2016perceptual, shih2013data, galteri2017deep, luc2017predicting, ledig2017photo}. Most approaches work with RGB images, typically augmenting or restoring them. Notable examples are super-resolution~\cite{ledig2017photo}, artifact removal~\cite{galteri2017deep}, style transfer~\cite{johnson2016perceptual} and multiple time of day generation~\cite{shih2013data}. On the other hand multimodal translations are possible, where RGB images are translated to edges~\cite{xie2015holistically}, depth and surfaces~\cite{eigen2015predicting} or segmentation maps~\cite{long2015fully, eigen2015predicting}. Recent image-to-image techniques based on Generative Adversarial Networks (GAN)~\cite{goodfellow2014generative}, used in a conditional setting, have provided more flexible architectures capable of addressing many translation tasks~\cite{isola2017image, CycleGAN2017} such as semantic segmentation to RGB, season change, sketch to RGB, aerial images to maps and style transfer.

Image inpainting~\cite{bertalmio2000image, yu2018generative, yeh2017semantic, pathak2016context, liu2018image}, refers to the task of predicting missing or damaged parts of an image, by inferring them from contextual information. Put in the framework of image transformation, inpainting can be seen as a special case of image-to-image translation with an additional constraint on where to restore the image. Inpainting has a large variety of applications, spanning from image editing and restoration to the more complex semantic inpainting \cite{pathak2016context}, where large image crops are reconstructed thanks to high level semantics of scene and objects. A recent trend has seen GANs as the main protagonists of image inpainting~\cite{yu2018generative, yeh2017semantic}, however existing methods focus on completing natural scene images and are limited to RGB images. 

In the automotive scenario, image generation has been mainly used as an augmentation procedure\cite{qi2018semi, wang2018vid2vid}. Nonetheless one of the main component of an autonomous driving system is trajectory planning and estimation~\cite{paden2016survey,franke2017autonomous}. For a proper motion planning, agents must know their surroundings. This is often obtained through a combination of feature based localization and 3D point-cloud registration~\cite{wolcott2017robust}. Knowledge of surroundings is mandatory to obtain physical constraints to be added to the trajectory prediction and planning problem.

 In this work we tackle the novel problem of semantic image inpainting, in which source images are obtained by an automatic algorithm~\cite{chen2018deeplab}. Semantic segmentation has only been used in the task of  inpainting~\cite{song2018spg}  to guide the RGB generation and obtain more pleasant reconstructions to the human eye. On the contrary, we completely discard the RGB content and focus on the semantics to reconstruct the signal hidden in the image itself, rather than its texture. The motivation rises from the need to precisely comprehend the structure of what is occluded when appearance is of relatively low importance compared to raw structure.
This is of particular interest for autonomous driving where clutter and occlusion are frequent, posing a threat to safety. Specifically, we want to understand the static layout of the scene by removing dynamic objects. Once such a layout is recovered it is possible to derive physical constraints from the scene that can be used in all reasoning tasks regarding own and other behaviors, such as path planning and more effective obstacle avoidance.

To the best of our knowledge, we are the first to propose a segmentation inpainting method to reconstruct the hidden semantics of a scene using GANs.
Despite being different in spirit, the closest approach to ours is~\cite{luc2017predicting} which casts the problem of future anticipation in automotive as a segmentation-to-segmentation task. However, apart from not performing inpainting, they focus on moving objects, while we want to recover static components of the environment.

The advantages of using semantic segmentations rather than RGB are twofold: on the one hand it allows us to identify dynamic objects at a pixel-level and localize occlusion; on the other hand it directly yields a complete understanding of the image.
Moreover, RGB inpainting methods still provide images which may be imprecise and of difficult interpretation. Our method instead is capable of inpainting directly the category of the restored pixels, excluding any uncertainty in the reconstruction.

\begin{figure*}
	\centering
	\begin{tabular}{ccc}
		Input  & Ours & Nearest Neighbor \\
		\csQualitative{0.3}{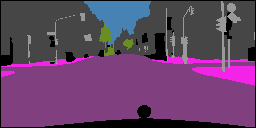} 
		\csQualitative{0.3}{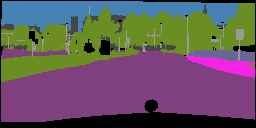}
		\csQualitative{0.3}{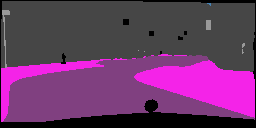}
		\csQualitative{0.3}{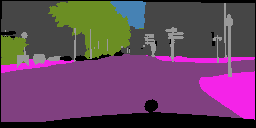}
		\csQualitative{0.3}{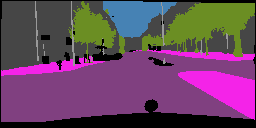}
		\csQualitative{0.3}{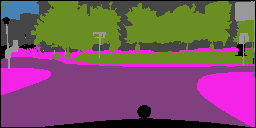}
		\csQualitative{0.3}{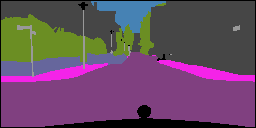}
		\csQualitative{0.3}{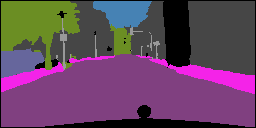}
		\csQualitative{0.3}{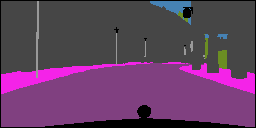}
		\csQualitative{0.3}{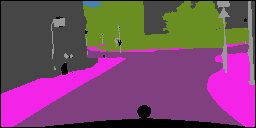}
	\end{tabular}
	\caption{Qualitative results on the Cityscapes dataset. NN inpainting is shown as comparison.}
	\label{img:cityscapes}
\end{figure*}

\section{Semantic Segmentation Inpainting}
In this paper we propose to inpaint dynamic objects in semantic segmentations of ego-vehicle images to recover the static road layout.
Given a set of visual categories $\mathcal{C} = \mathcal{S} \cup \mathcal{D}$, composed by a subset of static and dynamic classes $\mathcal{S}$ and $\mathcal{D}$, we convert a segmentation $\mathbf{I}$ with values in $\mathcal{C}$, into $\mathbf{O}$ with values in $\mathcal{S}$.
In Fig.~\ref{img:inpainting} are shown examples of inputs and outputs of our method, along with the binary masks that guide the inpainting of dynamic objects.

Our proposed model follows a Generative Adversarial Network paradigm: a generator is trained to generate plausible inpaintings and fool a discriminator, which is trained to recognize whether an image belongs to the real or reconstructed distributions of data. In this paper we modify this architecture to work with N-dimensional data instead of just RGB images. The input segmentation mask is fed to the network as a one-hot encoded tensor $\mathbf{I} \in \{0, 1\}^{W \times H \times \left | \mathcal{C} \right |}$ of the class labels in the semantic map, where $W$ is the width of the image, $H$ its height and $\left | \mathcal{C} \right |$ the number of classes. In the same way we train the network to output a new tensor $\mathbf{O} \in \{0, 1\}^{W \times H \times \left | \mathcal{S} \right |}$ with the same width and height but containing only categories belonging to the static set $\mathcal{S}$.

We extend the Generative Inpainting Network of \cite{yu2018generative}, where a coarse-to-fine approach is followed to generate RGB images. To adapt the network to a segmentation inpainting task, we changed the $\ell_1$ pixel-wise reconstruction loss to a softmax cross-entropy loss, casting the problem as a classification task instead of a regression one. This choice makes every class independent and forces a hard class assignment on the output, opposed to classical inpainting scenarios where a perceptually close RGB value is acceptable.

The authors of \cite{yu2018generative} introduced an attention layer to transform contextual regions into convolutional filters and estimate the correlation between background and foreground patches. This contextual attention is used to learn where to borrow image content and use it to guide the inpainting process. Since the reconstruction involves a certain degree of uncertainty, the model is trained with a spatially discounted loss, which avoids to penalize pixels far from the boundaries of the region to inpaint.

 Both stages of the generator proposed by Yu \textit{et al.}  \cite{yu2018generative} are constituted by 17 convolutional layers: 6 standard convolutional layers, with downsampling, are used first, then 4 atrous convolutional layers followed by two standard ones with a final upsampling block of 5 layers. Attention is used only in the second stage generator right after the atrous convolutional block.



We train our model using the manually annotated semantic segmentations of the Cityscapes dataset~\cite{cordts2016cityscapes}. For each image we consider a $256 \times 128 \times \left | \mathcal{C} \right |$ crop and randomly sample a rectangular binary mask of maximum size $64 \times 64$ within it. The portion of the input covered by the mask is then blacked out and reconstructed by the generator. The discriminator is fed with both original and reconstructed patches and trained to discriminate between them.


\section{Experimental Evaluation}
We trained our model on Cityscapes~\cite{cordts2016cityscapes}, an urban driving dataset with 30 pixel-wise annotated categories.
We have chosen the Cityscapes dataset since it contains a high variability of both static and dynamic categories and can therefore be adapted also to datasets comprising less categories.
In our experiments we divided the classes into the dynamic and static subcategories, clustering together similar ones.
The resulting 12 categories are the following.

\begin{equation*}
	\mathcal{D} = \{ \textit{Person},~ \textit{Car},~ \textit{Truck and Bus},~ \textit{Two Wheeled Vehicle} \}
\end{equation*}
\begin{equation*}
\mathcal{S} = \{ \textit{Road},~ \textit{Sidewalk and Parking},~ \textit{Building},~ \textit{Wall and Fence},~ \textit{Traffic Sign},
\end{equation*}
\begin{equation*} \textit{Vegetation and Terrain},~ \textit{Sky},~ \textit{Unlabeled} \}
\end{equation*}

At test time, we mask out all pixels belonging to dynamic classes and process the whole image in order to remove these objects.
As a source of data for the semantic segmentations, we use both at training and testing time the manually annotated segmentations provided with the dataset. A qualitative evaluation on the Cityscapes dataset is shown in Fig.~\ref{img:cityscapes}. Nonetheless we show how our method can also be applied to outputs of semantic segmentation methods such as Deeplab~\cite{chen2018deeplab} in Section~\ref{exp_deeplab}.

Images captured in the real world such as the Cityscapes dataset are hardly paired with exact copies of the scene that do not include occluding dynamic objects. This would require to collect pictures from the exact position and viewpoint of the original images, when no moving object is present. Moreover, to obtain a perfect pixel-wise alignment of the two images would require an image registration algorithm which would possibly lead to noise and empty regions. Due to this limitation, quantitatively assessing the quality of our inpainting model compared to other methods is not possible on natural images.

\begin{figure}[t]
\centering
\includegraphics[width=\textwidth]{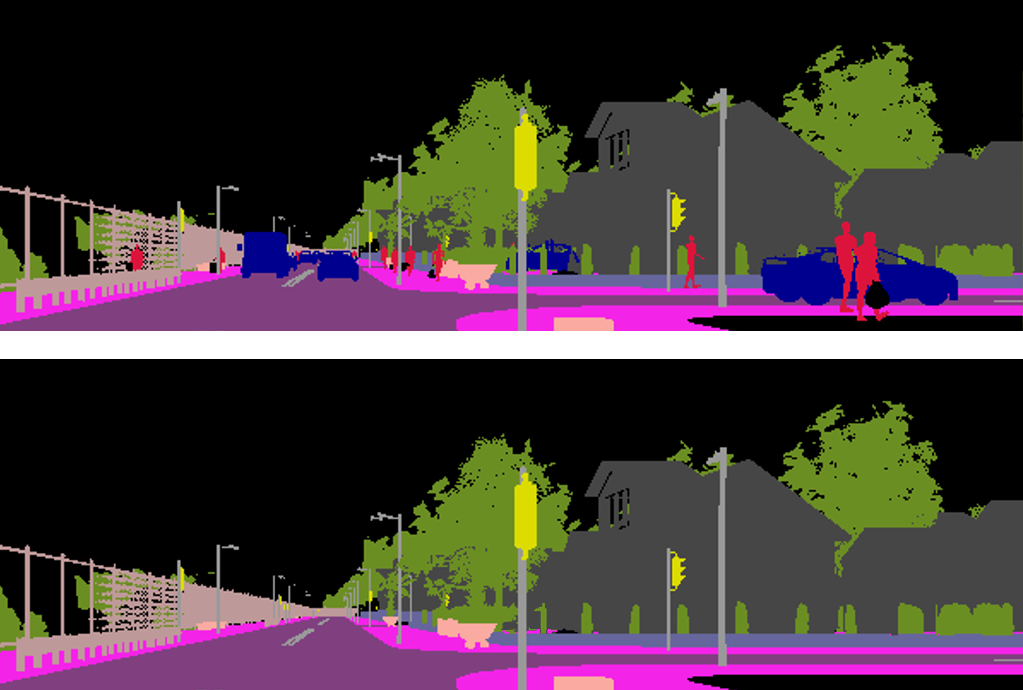}
\caption{In the MICC-SRI dataset we collected pixel-wise aligned segmentations with and without dynamic objects.}
\label{img:micc-sri}
\end{figure}

\subsection{MICC-SRI Semantic Road Inpatining dataset}

To overcome the quantitative evaluation problem on real world images, we generated an auxiliary dataset using CARLA~\cite{dosovitskiy2017carla}, an open-source urban driving simulator built under the Unreal Engine. Apart from providing a sandbox for autonomous driving algorithms, it offers functionalities for recording sequences varying the number of dynamic objects such as cars, pedestrians and two wheeled vehicles. The sequences can be acquired as almost photo-realistic RGB videos or converted on the fly into depth or semantic segmentation maps. Thanks to this functionality we are able to programmatically generate perfectly aligned pairs of pixel-wise semantic maps with and without dynamic objects.
This allows us to produce a ground truth reconstruction that would be impossible to obtain from real world images. Whereas CARLA generated RGB images are obviously distinguishable from natural images, semantic segmentations are instead very close to data acquired from real urban scenarios thanks to the lack of texture. The obtained pairs can therefore be used for quantitatively evaluating the models on the inpainting task.

In our work we used CARLA 0.8.2 which includes only cars and pedestrians as moving vehicles. Since release 0.8.3, the simulator also includes two wheeled vehicles such as bikes and motorbikes, but are reportedly still unstable and therefore we did not include them in our dataset.

To collect data we used autopilot simulations with CARLA in both its maps (Town01 and Town02), starting from all of their spawning points. The two maps have respectively 152 and 83 spawning points and for each simulation we gathered 1000 frames ran at 3 FPS (the minimum available rate) to increase variability, for a total of approximately 22 hours of driving simulation. The data has then been sampled at 0.3 FPS to remove redundancies. Note that the data acquisition process time has a 1:1 dependency with the simulation time.

All the simulations are run twice, once with dynamic objects and once without. To be able to obtain two matching versions of the same footage in both modalities we first ran the autopilot simulation with the map populated with dynamic objects and serializing all the commands given by the autopilot. In the second empty map simulation, instead on relying on the autopilot we load the driving commands previously acquired and make the ego vehicle follow the same exact path as before. For both simulations we save automatically generated semantic segmentations and RGB frames for reference, both at a resolution of 800x600 pixels.
 
Since CARLA is not fully deterministic we experienced a slight drift of the vehicle position with respect to the two versions of the same simulation. This drift appears to be triggered by rare events such as stopping and starting the car or by nondeterministic behaviors when getting close to other objects' colliders in the game engine. This drift becomes significant after 1000 frames, which is why we are collecting multiple short simulations rather than a few long runs. To correct the small misalignment due to this issue, we look for unmatched pixels belonging to static classes in the images with dynamic objects and replace them with the correct class. In order to remove trivial images with empty roads we consider only frames where the number of pixels belonging to moving objects is higher than 5000, i.e. approximately $0.001\%$ of the image.

This data acquisition process led to 11,913 pairs of perfectly aligned frames with and without dynamic objects. We refer to this novel dataset as MICC Semantic Road Inpainting dataset (MICC-SRI) and we released it for download at \url{www.micc.unifi.it/resources/datasets/semantic-road-inpainting/}. A sample of paired semantic segmentations from our dataset is shown in Fig.~\ref{img:micc-sri}


Thanks to this correspondence between occluded and non occluded pixels in the two versions of the image, we are now able to evaluate our method. We report per-pixel accuracy on the MICC-SRI dataset.

Since CARLA provides labels for fewer classes than CityScapes, we group together similar classes and remove the ones that are not present. The new set of classes used for experiments on the MICC-SRI dataset are the following:

\begin{equation*}
\mathcal{D} = \{ \textit{Person},~ \textit{Car} \}
\end{equation*}
\begin{equation*}
\mathcal{S} = \{ \textit{Road},~ \textit{Sidewalk},~ \textit{Building},~ \textit{Fence},~ \textit{Pole},\textit{Vegetation},~ \textit{Unlabeled} \}
\end{equation*}


\subsection{Baselines}

We propose several baselines, all performing both inpainting in the RGB and segmentation domain. Since to the best of our knowledge we are the first to perform inpainting in the semantic segmentation domain, we aim at demonstrating that traditional approaches for inpainting are not well suited for working directly with segmentations. This is due to the fact that inpainting methods often rely on image traits inside semantically correlated image regions such as textures or gradients. This information though is lost outside of the RGB domain.
The baselines we propose are the following:

\textbf{Nearest Neighbor (NN)}: for each masked pixel we retrieve the spatially closest pixel belonging to a static class and simply assign its class. In the presence of isolated small objects or uniform background this method can be quite effective to recover the rough geometry of the scene. On the other hand it is likely to fail when the scene is crowded and with complex backgrounds, especially when many region boundaries are occluded at the same time. We also inspect its variant in the RGB domain, transferring the color value instead of the class label among pixels. This approach is much less reliable due to the high variability of pixel values, leading to noisy reconstructions.

\textbf{Navier-Stokes~\cite{bertalmio2001navier}}: initially proposed as an RGB inpainting method, the Navier-Stokes approach is based on differential equations of fluid dynamics, posing an analogy between pixel intensities and two dimensional fluid stream functions. It follows the edges of known portions of the image up to the inpainted region and extends isophotes, i.e. edges with the same intensity, by matching gradient vectors on the inpainted region boundaries. Once the edges are connected from one part to another of the region, the internal pixels are filled in order to minimize variance within the area they belong to. The choice of this method was dictated by its nature of following contours rather than textures, which well adapts to semantic segmentations.

\textbf{PatchMatch \cite{barnes2009patchmatch}}: the algorithm establishes correspondences between patches in the image and attempts to replace the inpainted patch with the most relevant one. The matching process is based on a randomized algorithm to approximate Nearest Neighbors between two patches. After a random initial guess for the correspondence, the algorithm alternates between propagating good correspondences to close patches and sampling the neighboring regions. For the experiments using the PatchMatch algorithm we adopted the implementation available in the content-aware feature of Photoshop CC combined with its scripting functionalities.

\begin{table}[t]
	\centering
	\begin{tabular}{ccc|cccc}
		
		\multicolumn{1}{l|}{\multirow{2}{*}{~~~Input~~~}} & \multicolumn{1}{c|}{\multirow{2}{*}{~~~Pre-process~~~}} & \multicolumn{1}{c|}{\multirow{2}{*}{~~~Mask source~~~}} & \multicolumn{4}{c}{Inpainting Method}                     \\ 
		\multicolumn{1}{l|}{}                       & \multicolumn{1}{l|}{}                             & \multicolumn{1}{l|}{}                             & ~~~~NN~~~~    & ~~~~NS\cite{bertalmio2001navier}~~~~ & ~~~~PM\cite{barnes2009patchmatch}~~~~ & ~~~~GAN~~~~ \\ \hline \hline
		\multicolumn{1}{c|}{GT}                & \multicolumn{1}{c|}{-}                           & GT  & 68.41 & 29.18         & 19.32      & \textbf{81.94}        \\ \hline
		\multicolumn{1}{c|}{RGB}                   & \multicolumn{1}{c|}{DeepLab}                     & DeepLab                                           & 61.41 & 30.41         & 21.33      & \textbf{70.58}        \\
		\multicolumn{1}{c|}{RGB}                   & \multicolumn{1}{c|}{-}                           & DeepLab                                           & 38.04 & 51.97         & 63.82      & 59.77                
	\end{tabular}
	\caption{\label{tab:acc}Per-pixel accuracy on the MICC-SRI dataset. Each row is relative to a different processing pipeline of our method, as shown in the three models of Fig.~\ref{img:pipelines}. For each variant, the inpainting module can be performed by our method (GAN) or one of the proposed baselines: Nearest Neighbor (NN), Navier-Stokes~\cite{bertalmio2001navier} (NS) and PatchMatch~\cite{barnes2009patchmatch} (PM).}
\end{table}

\begin{figure}
	\begin{subfigure}[t]{\linewidth}
		\centering\includegraphics[height=0.3\textwidth]{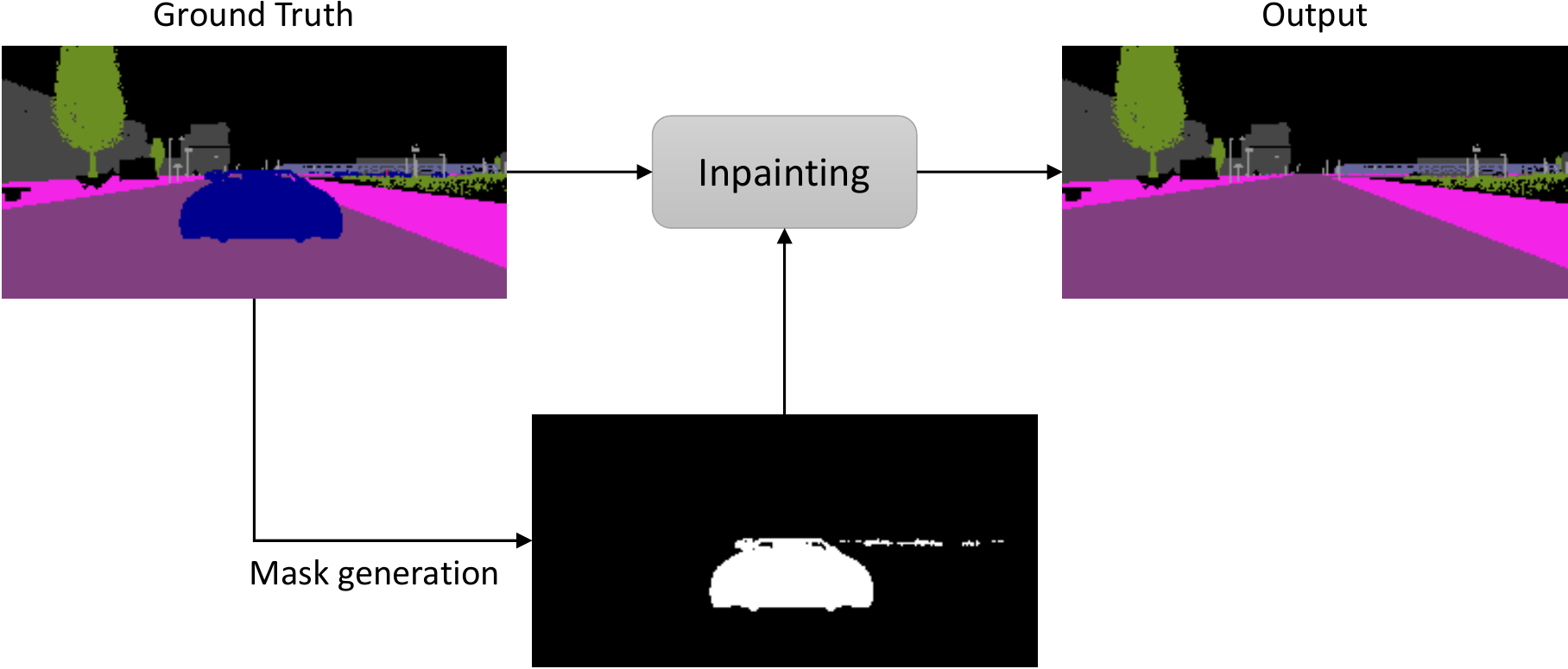}
		\caption{\textbf{Ground truth model}: the ground truth segmentation is fed to the inpainting module using pixels belonging to dynamic classes as the inpainting mask.\bigskip\bigskip}
		\label{img:pipelineA}
	\end{subfigure}
	\\
	\begin{subfigure}[t]{\linewidth}
		\centering\includegraphics[height=0.3\textwidth]{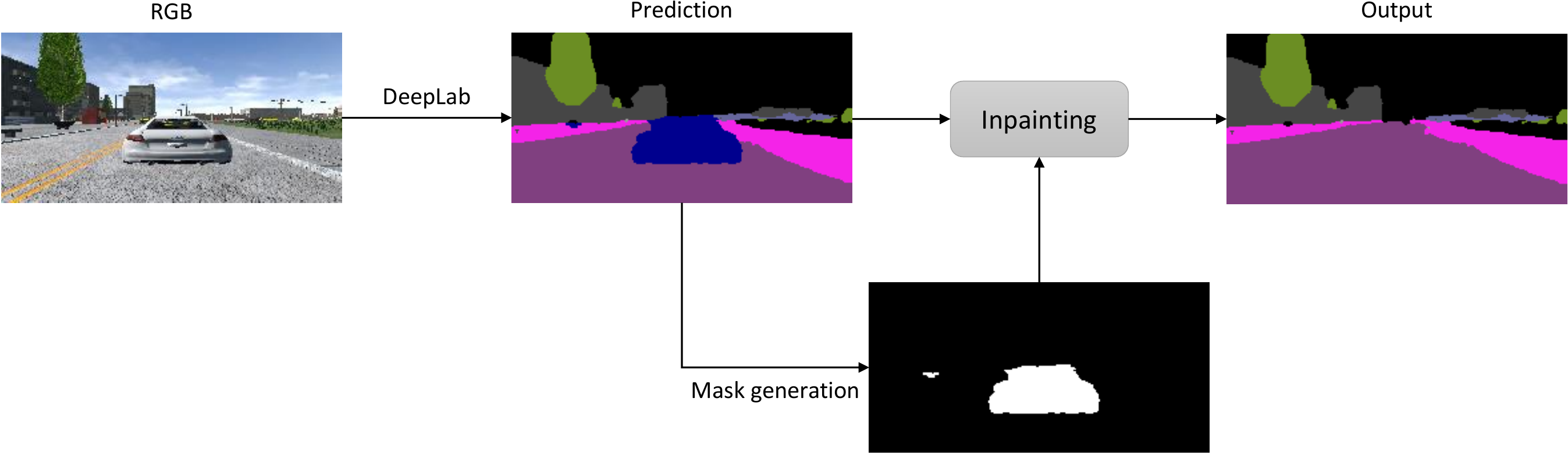}
		\caption{\textbf{Segmentation model}: the input segmentation is obtained using DeepLab~\cite{chen2017rethinking}. The inpainting mask is obtained from DeepLab's predictions.\bigskip\bigskip}
		\label{img:pipelineB}
	\end{subfigure}
	\\
	\begin{subfigure}[t]{\linewidth}
		\centering\includegraphics[height=0.3\textwidth]{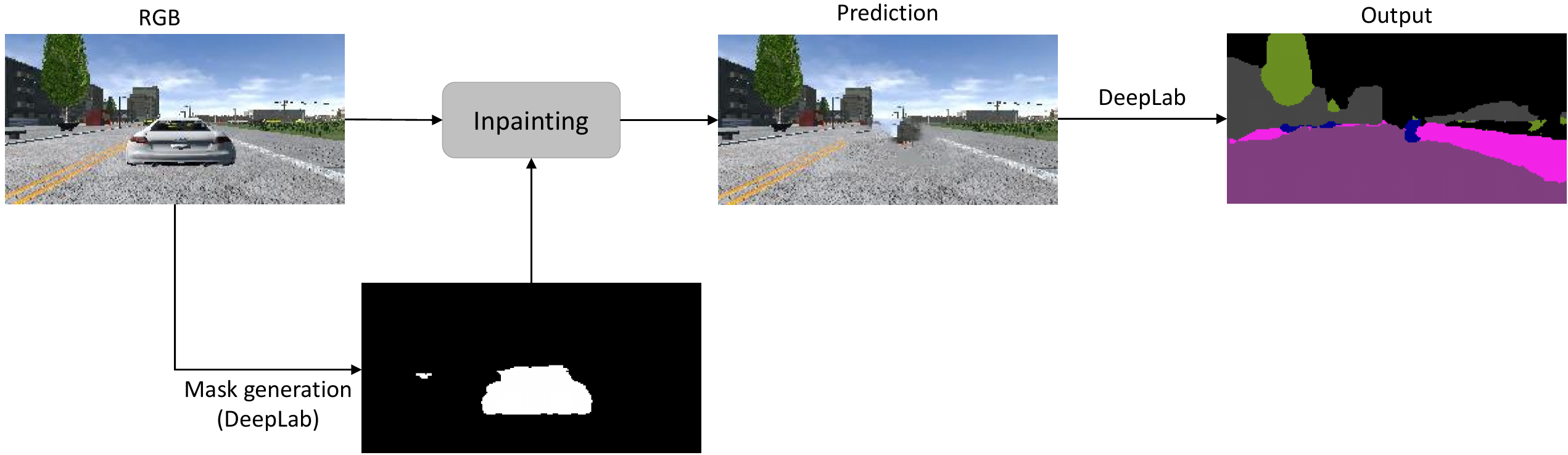}
		\caption{\textbf{RGB model}: the inpainting is performed on the RGB before applying DeepLab to get the final segmentation. A preliminary segmentation must be obtained to know where to inpaint.}
		\label{img:pipelineC}
	\end{subfigure}
	\caption{Pipeline of the variants of our architecture. Note that the inpainting step can be performed either by our model or by one of the proposed baselines.}
	\label{img:pipelines}
\end{figure}

\subsection{Ground Truth segmentations}

\begin{figure*}[]
	\centering
	\begin{tabular}{ccc}
		\MICCsriQualitativeBaselines{0.3}{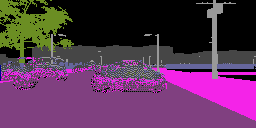}
		\MICCsriQualitativeBaselines{0.3}{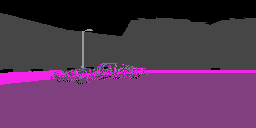}
		\MICCsriQualitativeBaselines{0.3}{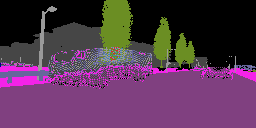}
		\MICCsriQualitativeBaselines{0.3}{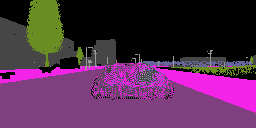}
		\MICCsriQualitativeBaselines{0.3}{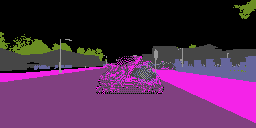}
	\end{tabular}
	\caption{MICC-SRI qualitative results. Ground Truth segmentations are used as input.}
	\label{img:carla_qualitative}
\end{figure*}

The first row of Tab.~\ref{tab:acc} shows a comparison of our method against all the proposed baselines on the MICC-SRI dataset. Here we use as input the ground truth automatically acquired with the CARLA simulator. All pixels belonging to dynamic classes are used to mask the image and inpaint it in one single step. A schematic representation of our model is depicted in Fig.~\ref{img:pipelineA}. We report pixel wise accuracy measured within the inpainted mask, i.e. the percentage of correctly inpainted pixels in each mask averaged across images.
Our method outperforms by a large margin all the baselines, proving that RGB inpainting methods are not suitable for images without textures. Nearest-Neighbor performs reasonably well compared to the other baselines, yet fails to grasp the layout of the scene since no reasoning is involved in the generation process. To better understand strengths and flaws of all the methods, in Fig.~\ref{img:carla_qualitative} a qualitative comparison is given. 

Nearest-Neighbor tends to hallucinate roads in sidewalks when removing cars and heavily distorts the overall layout at the horizon. Navier-Stokes manages to join edges across the inpainting masks but at the same time the reconstructions it provides are extremely noisy, adding noise patterns to the filled regions. If this is acceptable in RGB images, noise in semantic segmentations translates to a misclassification of pixels and therefore a lack of understanding of the scene.

The quality of results for the PatchMatch baseline may vary a lot depending on the input image. When the layout is simple and the algorithm can easily establish correspondences with the background, the reconstructions are reasonable. Instead when there are too many structural elements in the scene the algorithm tends to copy them into unnatural positions, for instance adding trees and traffic signs in the middle of the road. Again, this is a behavior that is suitable for highly textured regions, whereas flat regions as semantic segmentations should just be filled with an uniform pattern depending on the expected layout.

\begin{figure*}[]
	\centering
	\includegraphics[width=0.7\textwidth]{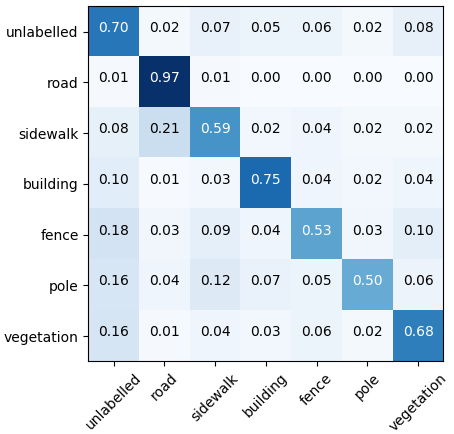}
	\caption{Class accuracy confusion matrix for our method over GT segmentations.\\True labels on the y-axis, predicted ones on the x-axis}
	\label{img:confmat_gt}
\end{figure*}

In Fig.~\ref{img:confmat_gt} we report the confusion matrix for our method to provide insights on how it is performing. Interestingly the \textit{Road} class is almost perfectly reconstructed, but at the same time the \textit{Sidewalk} class is sometimes confused with \textit{Road}, probably due to close proximity in the data.
The \textit{Unlabeled} class is, on average, the most common among the wrongly-predicted classes which is a reasonable outcome given that \textit{Unlabeled} is used as a catch-all class for everything that does not fit in the other classes, such as the sky, small objects, and urban design elements.

\begin{figure*}[]
	\centering
	\begin{tabular}{ccc}
		\MICCexpTwo{0.3}{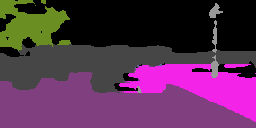}
		\MICCexpTwo{0.3}{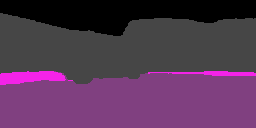}
		\MICCexpTwo{0.3}{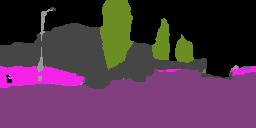}
		\MICCexpTwo{0.3}{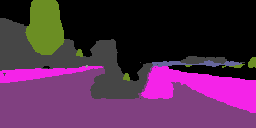}
		\MICCexpTwo{0.3}{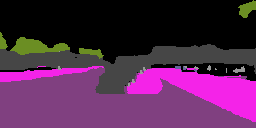}
	\end{tabular}
	\caption{MICC-SRI qualitative results. DeepLab segmentations are used as input.}
	\label{img:carla_qualitative_deeplab}
\end{figure*}

\begin{figure*}[]
	\centering
	\begin{tabular}{ccc}
		\MICCexpThree{0.3}{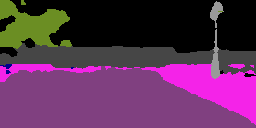}
		\MICCexpThree{0.3}{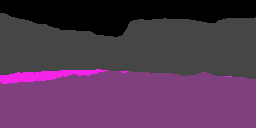}
		\MICCexpThree{0.3}{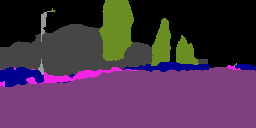}
		\MICCexpThree{0.3}{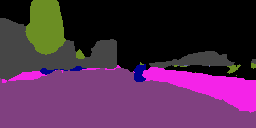}
		\MICCexpThree{0.3}{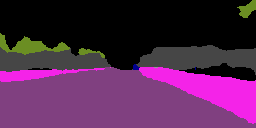}
	\end{tabular}
	\caption{MICC-SRI qualitative results. The inpainting is done on RGB image and the output is then segmented using DeepLab.}
	\label{img:carla_qualitative_rgb}
\end{figure*}


\subsection{Predicted segmentations}
\label{exp_deeplab}
Our method is made to aid autonomous driving agents, which will require to understand the scene as they are deployed on the street. Therefore in a real case scenario, our method cannot rely on manually annotated or automatically generated semantic segmentations and an alternative source will have to be exploited. Many semantic segmentation algorithms have been proposed in literature \cite{ronneberger2015u, badrinarayanan2015segnet, long2015fully, paszke2016enet, chen2018deeplab, chen2017rethinking}, often with a special attention to autonomous driving applications. Here we show how our method can be also effectively applied on generated semantic segmentations as source instead of ground truth annotations. In our experiments we use predictions from DeepLab-v3+~\cite{chen2017rethinking}, which we trained on segmentations generated with CARLA to obtain a compatible mapping with the categories we want to predict. Experimental results are reported in Tab.~\ref{tab:acc}.

We report two variants of our experiments, depending on the order in which we apply our building blocks. In one case we first apply DeepLab on the RGB image to obtain the segmentations and then we inpaint the resulting map to remove pixels classified as dynamic classes. This model is shown in Fig.~\ref{img:pipelineB}. In the other case we perform the inpainting step in the RGB domain and then we apply DeepLab on the inpainted image. The RGB inpainting requires two segmentation steps: the first one is applied on the RGB source and is needed to localize dynamic object pixels and create an inpainting mask, the second one is applied on the inpainted RGB and is necessary to obtain the final segmentation output. The pipeline for this variant of the model is shown in Fig.~\ref{img:pipelineC}.

Results on the MICC-SRI dataset for the segmentation and RGB inpainting pipelines are reported in Tab.~\ref{tab:acc} in the second and third row, respectively.
When replacing ground truth segmentations with automatically generated segmentations, our method still performs better than the baselines reconstructing reasonable layouts. Segmented regions produced by DeepLab tend to exhibit smoother boundaries than the ground truth, often turning straight contours into noisy and curved lines. This makes the inpainting task harder since less natural boundaries are more difficult to follow and join together over the inpainted region. To evaluate the accuracy in this case we adopt as ground truth the outputs generated by DeepLab on the static version of the frames. This introduces a further level of uncertainty in the evaluation due to the aforementioned fluctuations of region boundaries. 
With Navier-Stokes and PatchMatch instead the usual pathological behaviors are present, keeping the results low as in the ground truth version. Qualitative results are shown in Fig.~\ref{img:carla_qualitative_deeplab}.

The opposite trend can be observed when the inpainting is made on RGB images and DeepLab is applied on the resulting image. In this case instead of our method for inpainting images we use the original formulation of the Generative Inpainting model of Yu \textit{et al.}~\cite{yu2018generative}. All the RGB inpainting methods perform roughly on par, with the exception of Nearest Neighbor which drops 20 points below the others since it generates unnatural reconstructions. Overall though, RGB methods provide much lower accuracy in the final segmentations compared to our semantic inpainting model. Qualitative results are shown in Fig.~\ref{img:carla_qualitative_rgb}.

\begin{figure*}[]
	\centering
	\includegraphics[width=0.7\textwidth]{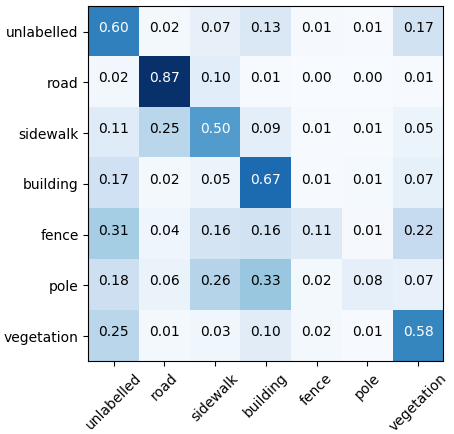}
	\caption{Class accuracy confusion matrix for our method over DeepLab segmentations. True labels on the y-axis, predicted ones on the x-axis}
	\label{img:confmat_deeplab}
\end{figure*}

We report the confusion matrix for our method over DeepLab segmentations in Fig.~\ref{img:confmat_deeplab}. Here, similarly to  Fig.~\ref{img:confmat_gt}, the \textit{Road} class performs well, and so do the other classes that are easily picked up by DeepLab due to their size in the image (\textit{Building, Sidewalk, Unlabelled}).
The accuracy of classes that have finer, smaller objects (\textit{Fence, Pole}) drops dramatically since the DeepLab segmentation fails over them.

\section{Conclusions}
In this paper we presented a segmentation-to-segmentation inpainting model to recover the layout of an urban driving scenario. To the best of our knowledge we are the first to propose a model for inpainting semantic segmentations.
The model we presented is a Generative Adversarial Network architecture capable of removing dynamic objects from the scene and reconstruct occluded views. We showed the effectiveness of the model both on Cityscapes and on a novel synthetically generated dataset obtained with the CARLA simulator and freely available online. Along with our method we presented several baselines working both in the RGB and the segmentation domain. The comparisons of the different methods highlighted the benefits of working directly with segmentations rather than segmenting inpainted images. We also showed that classic RGB inpainting methods are not suitable when working outside from the highly structured and textured domain of natural images. We believe that being able to infer occluded regions in autonomous driving systems is a key component to achieve a full comprehension of the scene and will allow better planning of the ego-vehicle trajectories in crowded urban scenarios.

\bigskip

\textbf{Acknowledgements} We gratefully acknowledge the support of NVIDIA Corporation with the donation of the Titan Xp GPU used for this research.

\bibliographystyle{spmpsci}
\bibliography{egbib}

\begin{thebibliography}{10}
\providecommand{\url}[1]{{#1}}
\providecommand{\urlprefix}{URL }
\expandafter\ifx\csname urlstyle\endcsname\relax
  \providecommand{\doi}[1]{DOI~\discretionary{}{}{}#1}\else
  \providecommand{\doi}{DOI~\discretionary{}{}{}\begingroup
  \urlstyle{rm}\Url}\fi

\bibitem{badrinarayanan2015segnet}
Badrinarayanan, V., Kendall, A., Cipolla, R.: Segnet: A deep convolutional
  encoder-decoder architecture for image segmentation.
\newblock arXiv preprint arXiv:1511.00561  (2015)

\bibitem{barnes2009patchmatch}
Barnes, C., Shechtman, E., Finkelstein, A., Goldman, D.B.: Patchmatch: A
  randomized correspondence algorithm for structural image editing.
\newblock ACM Transactions on Graphics (ToG) \textbf{28}(3), 24 (2009)

\bibitem{bertalmio2001navier}
Bertalmio, M., Bertozzi, A.L., Sapiro, G.: Navier-stokes, fluid dynamics, and
  image and video inpainting.
\newblock In: Computer Vision and Pattern Recognition, 2001. CVPR 2001.
  Proceedings of the 2001 IEEE Computer Society Conference on, vol.~1, pp.
  I--I. IEEE (2001)

\bibitem{bertalmio2000image}
Bertalmio, M., Sapiro, G., Caselles, V., Ballester, C.: Image inpainting.
\newblock In: Proceedings of the 27th annual conference on Computer graphics
  and interactive techniques, pp. 417--424. ACM Press/Addison-Wesley Publishing
  Co. (2000)

\bibitem{chen2018deeplab}
Chen, L.C., Papandreou, G., Kokkinos, I., Murphy, K., Yuille, A.L.: Deeplab:
  Semantic image segmentation with deep convolutional nets, atrous convolution,
  and fully connected crfs.
\newblock IEEE transactions on pattern analysis and machine intelligence
  \textbf{40}(4), 834--848 (2018)

\bibitem{chen2017rethinking}
Chen, L.C., Papandreou, G., Schroff, F., Adam, H.: Rethinking atrous
  convolution for semantic image segmentation.
\newblock arXiv preprint arXiv:1706.05587  (2017)

\bibitem{cordts2016cityscapes}
Cordts, M., Omran, M., Ramos, S., Rehfeld, T., Enzweiler, M., Benenson, R.,
  Franke, U., Roth, S., Schiele, B.: The cityscapes dataset for semantic urban
  scene understanding.
\newblock In: Proceedings of the IEEE conference on computer vision and pattern
  recognition, pp. 3213--3223 (2016)

\bibitem{dosovitskiy2017carla}
Dosovitskiy, A., Ros, G., Codevilla, F., L{\'o}pez, A., Koltun, V.: Carla: An
  open urban driving simulator.
\newblock arXiv preprint arXiv:1711.03938  (2017)

\bibitem{eigen2015predicting}
Eigen, D., Fergus, R.: Predicting depth, surface normals and semantic labels
  with a common multi-scale convolutional architecture.
\newblock In: Proceedings of the IEEE International Conference on Computer
  Vision, pp. 2650--2658 (2015)

\bibitem{franke2017autonomous}
Franke, U.: Autonomous Driving, chap.~2, pp. 24--54.
\newblock Wiley-Blackwell (2017).
\newblock \doi{10.1002/9781118868065.ch2}.
\newblock
  \urlprefix\url{https://onlinelibrary.wiley.com/doi/abs/10.1002/9781118868065.ch2}

\bibitem{galteri2017deep}
Galteri, L., Seidenari, L., Bertini, M., Del~Bimbo, A.: Deep generative
  adversarial compression artifact removal.
\newblock arXiv preprint arXiv:1704.02518  (2017)

\bibitem{goodfellow2014generative}
Goodfellow, I., Pouget-Abadie, J., Mirza, M., Xu, B., Warde-Farley, D., Ozair,
  S., Courville, A., Bengio, Y.: Generative adversarial nets.
\newblock In: Advances in neural information processing systems, pp. 2672--2680
  (2014)

\bibitem{isola2017image}
Isola, P., Zhu, J.Y., Zhou, T., Efros, A.A.: Image-to-image translation with
  conditional adversarial networks.
\newblock arXiv preprint  (2017)

\bibitem{johnson2016perceptual}
Johnson, J., Alahi, A., Fei-Fei, L.: Perceptual losses for real-time style
  transfer and super-resolution.
\newblock In: European Conference on Computer Vision, pp. 694--711. Springer
  (2016)

\bibitem{kirillov2018panoptic}
Kirillov, A., He, K., Girshick, R., Rother, C., Doll{\'a}r, P.: Panoptic
  segmentation.
\newblock arXiv preprint arXiv:1801.00868  (2018)

\bibitem{ledig2017photo}
Ledig, C., Theis, L., Husz{\'a}r, F., Caballero, J., Cunningham, A., Acosta,
  A., Aitken, A.P., Tejani, A., Totz, J., Wang, Z., et~al.: Photo-realistic
  single image super-resolution using a generative adversarial network.
\newblock In: CVPR, vol.~2, p.~4 (2017)

\bibitem{liu2018image}
Liu, G., Reda, F.A., Shih, K.J., Wang, T.C., Tao, A., Catanzaro, B.: Image
  inpainting for irregular holes using partial convolutions.
\newblock arXiv preprint arXiv:1804.07723  (2018)

\bibitem{long2015fully}
Long, J., Shelhamer, E., Darrell, T.: Fully convolutional networks for semantic
  segmentation.
\newblock In: Proceedings of the IEEE conference on computer vision and pattern
  recognition, pp. 3431--3440 (2015)

\bibitem{luc2017predicting}
Luc, P., Neverova, N., Couprie, C., Verbeek, J., LeCun, Y.: Predicting deeper
  into the future of semantic segmentation.
\newblock In: IEEE International Conference on Computer Vision (ICCV), vol.~1
  (2017)

\bibitem{paden2016survey}
Paden, B., {\v{C}}{\'a}p, M., Yong, S.Z., Yershov, D., Frazzoli, E.: A survey
  of motion planning and control techniques for self-driving urban vehicles.
\newblock IEEE Transactions on intelligent vehicles \textbf{1}(1), 33--55
  (2016)

\bibitem{paszke2016enet}
Paszke, A., Chaurasia, A., Kim, S., Culurciello, E.: Enet: A deep neural
  network architecture for real-time semantic segmentation.
\newblock arXiv preprint arXiv:1606.02147  (2016)

\bibitem{pathak2016context}
Pathak, D., Krahenbuhl, P., Donahue, J., Darrell, T., Efros, A.A.: Context
  encoders: Feature learning by inpainting.
\newblock In: Proceedings of the IEEE Conference on Computer Vision and Pattern
  Recognition, pp. 2536--2544 (2016)

\bibitem{qi2018semi}
Qi, X., Chen, Q., Jia, J., Koltun, V.: Semi-parametric image synthesis.
\newblock In: Proceedings of the IEEE Conference on Computer Vision and Pattern
  Recognition, pp. 8808--8816 (2018)

\bibitem{richter2016playing}
Richter, S.R., Vineet, V., Roth, S., Koltun, V.: Playing for data: Ground truth
  from computer games.
\newblock In: European Conference on Computer Vision, pp. 102--118. Springer
  (2016)

\bibitem{ronneberger2015u}
Ronneberger, O., Fischer, P., Brox, T.: U-net: Convolutional networks for
  biomedical image segmentation.
\newblock In: International Conference on Medical image computing and
  computer-assisted intervention, pp. 234--241. Springer (2015)

\bibitem{shih2013data}
Shih, Y., Paris, S., Durand, F., Freeman, W.T.: Data-driven hallucination of
  different times of day from a single outdoor photo.
\newblock ACM Transactions on Graphics (TOG) \textbf{32}(6), 200 (2013)

\bibitem{song2018spg}
Song, Y., Yang, C., Shen, Y., Wang, P., Huang, Q., Kuo, C.C.J.: Spg-net:
  Segmentation prediction and guidance network for image inpainting.
\newblock arXiv preprint arXiv:1805.03356  (2018)

\bibitem{uhrig2016pixel}
Uhrig, J., Cordts, M., Franke, U., Brox, T.: Pixel-level encoding and depth
  layering for instance-level semantic labeling.
\newblock In: German Conference on Pattern Recognition, pp. 14--25. Springer
  (2016)

\bibitem{wang2018vid2vid}
Wang, T.C., Liu, M.Y., Zhu, J.Y., Liu, G., Tao, A., Kautz, J., Catanzaro, B.:
  Video-to-video synthesis.
\newblock arXiv preprint arXiv:1808.06601  (2018)

\bibitem{wang2018depth}
Wang, W., Neumann, U.: Depth-aware cnn for rgb-d segmentation.
\newblock arXiv preprint arXiv:1803.06791  (2018)

\bibitem{wolcott2017robust}
Wolcott, R.W., Eustice, R.M.: Robust lidar localization using multiresolution
  gaussian mixture maps for autonomous driving.
\newblock The International Journal of Robotics Research \textbf{36}(3),
  292--319 (2017)

\bibitem{xie2015holistically}
Xie, S., Tu, Z.: Holistically-nested edge detection.
\newblock In: Proceedings of the IEEE international conference on computer
  vision, pp. 1395--1403 (2015)

\bibitem{yeh2017semantic}
Yeh, R.A., Chen, C., Lim, T.Y., Schwing, A.G., Hasegawa-Johnson, M., Do, M.N.:
  Semantic image inpainting with deep generative models.
\newblock In: CVPR, vol.~2, p.~4 (2017)

\bibitem{yu2018generative}
Yu, J., Lin, Z., Yang, J., Shen, X., Lu, X., Huang, T.S.: Generative image
  inpainting with contextual attention.
\newblock arXiv preprint arXiv:1801.07892  (2018)

\bibitem{CycleGAN2017}
Zhu, J.Y., Park, T., Isola, P., Efros, A.A.: Unpaired image-to-image
  translation using cycle-consistent adversarial networks.
\newblock In: Computer Vision (ICCV), 2017 IEEE International Conference on
  (2017)

\end{thebibliography}
\end{document}